\documentclass[10pt]{article} % For LaTeX2e

\usepackage[accepted]{rlj} % Should be uncommented for the camera-ready
\usepackage{amssymb}            % Defines common symbols like \mathbb R
\usepackage{mathtools}          % Extends amsmath, providing common math tools
\usepackage{mathrsfs}           % Enables \mathscr, which can work in cases that \mathcal does not
\usepackage{graphicx}           % For including images
\usepackage{subcaption}         % Allows for the use of subfigures and subcaptions
\usepackage[space]{grffile}     % For spaces in image names
\usepackage{url}                % For displaying URLs
\usepackage{multirow}           % For multi-row cells in tables
\usepackage{booktabs}           % For \toprule, \midrule, \bottomrule in tables
\usepackage{lipsum}             % For placeholder text
\usepackage{algorithmic}
\usepackage{algorithm}

\usepackage{amsmath}
\usepackage{amssymb}
\usepackage{setspace}
\usepackage{listings}
\usepackage{xcolor}
\usepackage{courier}
\usepackage{relsize}

\title{Should We Ever Prefer Decision Transformer \\for Offline Reinforcement Learning?}

\setrunningtitle{Should We Ever Prefer Decision Transformer?}

\author{Yumi Omori\textsuperscript{1$\ast$}, Zixuan Dong\textsuperscript{2,3$\ast$}, Keith Ross\textsuperscript{1$\dagger$}}

\emails{keithwross@nyu.edu}

\affiliations{
$^{1}$\textbf{New York University Abu Dhabi}\\
$^{2}$\textbf{SFSC of AI and DL, NYU Shanghai}\\
$^{3}$\textbf{New York University}
\par % If including additional comments like below, use \par to add some whitespace. 
$^\ast$ Equal Contribution $^\dagger$ Corresponding author
}

\contribution{
    Provide a succinct but precise list of the contribution(s) of the paper. Use contextual notes to avoid implications of contributions more significant than intended and to clarify and situate the contribution relative to prior work (see the examples below). If there is no additional context, enter ``None''. Try to keep each contribution to a single sentence, although multiple sentences are allowed when necessary. If using complete sentences, include punctuation. If using a single sentence fragment, you may omit the concluding period. A single contribution can be sufficient, and there is no limit on the number of contributions. Submissions will be judged mostly on the contributions claimed on their cover pages and the evidence provided to support them. Major contributions should not be claimed in the main text if they do not appear on the cover page. Overclaiming can lead to a submission being rejected, so it is important to have well-scoped contribution statements on the cover page.
    }
    {
    None
    }

\contribution{
    The submission template for submissions to RLJ/RLC 2025
    }
    {
    Built from previous RLC/RLJ, ICLR, and TMLR submission templates
    }

\contribution{
    \textit{{[}Example of one contribution and corresponding contextual note for the paper ``Policy gradient methods for reinforcement learning with function approximation'' \citep{Sutton2000}.]}\\ This paper presents an expression for the policy gradient when using function approximation to represent the action-value function.
    }
    {
    Prior work established expressions for the policy gradient without function approximation \citep{Williams1992}.
    }

\keywords{RLJ, RLC, formatting guide, style file, \LaTeX~template.} % Your keywords

\summary{The summary appears on the cover page. Although it can be identical to the abstract, it does not have to be. One might choose to omit the stated contributions in the Summary, given that they will be stated in the box below. The original abstract may also be extended to two paragraphs. The authors should ensure that the contents of the cover page fit entirely on a single page. The cover page does \textbf{not} count towards the 8--12 page limit.

\lipsum[1]
}

\begin{document}

% \makeCover  % Create the cover page
\maketitle  % Make the title section

\begin{abstract}
% The transformer architecture, originally introduced for natural language processing, has rapidly become a foundational model across multiple domains, including a broad range of NLP and computer vision tasks. 

In recent years, extensive work has explored the application of the Transformer architecture to reinforcement learning problems. Among these, Decision Transformer (DT) \citep{Chen2021DecisionTransformer} has gained particular attention in the context of offline reinforcement learning due to its ability to frame return-conditioned policy learning as a sequence modeling task. Most recently, \citet{Bhargava2024WhenDT} provided a systematic comparison of DT with more conventional MLP-based offline RL algorithms, including Behavior Cloning (BC) and Conservative Q-Learning (CQL), and claimed DT exhibits superior performance in sparse-reward and low-quality data settings. 

In this paper, through experimentation on robotic manipulation tasks (Robomimic) and locomotion benchmarks (D4RL), we show that MLP-based Filtered Behavior Cloning (FBC) achieves competitive or superior performance compared to DT in sparse-reward environments. FBC simply filters out low-performing trajectories from the dataset and then performs ordinary behavior cloning on the filtered dataset. FBC is not only very straightforward, but it also requires less training data and is computationally more efficient. The results therefore suggest that DT is not preferable for sparse-reward environments. From prior work, arguably, DT is also not preferable for dense-reward environments. Thus, we pose the question: Is DT ever preferable?

% In this paper, we present a simple yet effective MLP-based algorithm for sparse-reward problems, Filtered Behavioral Cloning (FBC). FBC simply filters out low-performing trajectories from the dataset, and then performs ordinary behavioral cloning on the filtered dataset. Through extensive experimentation on robotic manipulation tasks (Robomimic) and locomotion benchmarks (D4RL), we demonstrate that FBC with  an MLP backbone achieves competitive or superior performance compared to DT in sparse-reward environments. Moreover, FBC has fewer parameters, requires less training data, and is computationally more efficient. Thus our results suggest that DT is not preferable for sparse-reward environments. From prior work, arguably DT is also not preferable for dense-reward environments. Thus we pose the question: Is DT ever preferable?

\end{abstract}

\section{Introduction}
The transformer architecture \citep{Vaswani2017Attention}, originally introduced for natural language processing, has rapidly become a foundational model across multiple domains, including a broad range of NLP and computer vision tasks. In recent years, there has been growing interest in the field of reinforcement learning (RL) in integrating transformers into the policy learning pipeline \citep{li2023DTsurvey}, notably Decision Transformer \citep{Chen2021DecisionTransformer}, and Trajectory Transformer \citep{Janner2021TrajectoryTransformer}. Among these, Decision Transformer (DT) has garnered particular attention in the context of offline reinforcement learning due to its ability to frame return-conditioned policy learning as a sequence modeling task. 

Recently, \citet{Bhargava2024WhenDT} provided a thorough comparison of DT with conventional MLP-based offline RL algorithms, including Behavior Cloning (BC) and Conservative Q-Learning (CQL) \citep{Kumar2020CQL}. They primarily examined two offline datasets: the D4RL \citep{fu2020d4rl} dataset and the Robomimic \citep{robosuiteDocs} dataset. For both datasets, they consider both dense reward and sparse reward variations. In the case of sparse rewards, through extensive experimentation, they argue that DT is preferable to conventional offline algorithms such as BC and CQL. For the dense-reward versions, their experiments show that CQL is preferable to DT for the D4RL dataset, and that DT is preferable to CQL for the Robomimic datasets. 

In this paper, we reexamine the question of whether DT is preferable for sparse-reward environments. To this end, we present a simple yet effective MLP-based algorithm, Filtered Behavior Cloning (FBC). FBC simply filters out low-performing trajectories from the dataset and then performs ordinary behavior cloning with the remaining trajectories.
On robotic manipulation tasks (Robomimic) and locomotion benchmarks (D4RL), we demonstrate that FBC with a small multi-layer perceptron (MLP) backbone achieves competitive or superior performance compared to DT in sparse-reward environments.
For example, for the D4RL datasets with sparsification, FBC beats DT for 7 of the 9 datasets, and improves aggregate performance by about 4\%. For the Robomimic dataset, FBC beats DT for both of the two datasets considered in \citet{Bhargava2024WhenDT}, and provides an aggregate performance improvement of about 3.5\%. Moreover, FBC uses fewer parameters, uses less training data, and learns more quickly in terms of wall-clock time. 
To make a fair comparison, all of the experiments reported in this paper follow closely the setups in \citet{Bhargava2024WhenDT}.

Additionally, we also consider Filtered Decision Transformer (FDT), where DT learns from the same filtered dataset as FBC does. Here we find that for the D4RL benchmarks, FBC performs better than FDT, and for the Robomimic benchmark, FBC and FDT perform about the same.

% \dzx{Rigorously, FDT is not FBC with transformed backbone. FDT still conditioned on RTG. For sparsified tasks, RTG are different for different trajs. Also, FDT considers history through attention.}\ym{when context = 1, i don't think they consider the history, but they do consider the relationship between state and action and RTG more. my original thinking was: In sparse and sparsified settings, rewards and therefore RTGs are binary, typically 1 for successful trajectories and 0 otherwise. As a result, RTG becomes an uninformative constant during autoregressive prediction, especially when the context length is fixed at 1. In such cases, DT effectively reduces to behavior cloning with a fixed RTG token. The only role RTG plays during training is to indicate that states paired with an RTG value of 0 are associated with failed trajectories. do you think this is correct? if so, perhaps revise it based on this line of thinking. }

Although we do not study dense-reward environments in this paper, based on prior literature \citep{Emmons2021Rvs, tarasov2023corl, yamagata2023QLearningDT, hu2024QRegularizedTransformer}, DT is arguably also not preferable to conventional MLP-based offline algorithms for dense-reward datasets in general. Thus, we pose the question: \emph{Is DT ever preferable for raw-state robotic tasks?}

\section{Related Work}
In addition to leveraging the attention mechanism of the Transformer to encode sequential observation histories as meaningful representations for downstream decision-making tasks \citep{tang2021sensory, guhur2023instruction, parisotto2020stabilizing, li2022pre}, the Transformer architecture itself is capable of formulating and solving decision-making problems as sequence modeling tasks. Some of the major examples include Decision Transformer (DT) \citep{Chen2021DecisionTransformer}, Trajectory Transformer (TT) \citep{Janner2021TrajectoryTransformer}, Generalized Decision Transformer (GDT) \citep{furuta2022GeneralizedDT}, Bootstrapped Transformer (BooT) \citep{wang2022bootstrapped}, Behavior Transformer (BeT) \citep{Shafiullah2022BehaviorTransformer}, and Q-Transformer \citep{chebotar2023QTransformer}. These extensions aim to better align transformer models with the structure of decision-making tasks. 

Decision Transformer, in particular, has received considerable attention to date with over 2000 citations as of June 2025. DT takes in as input sequences of states, actions, and return-to-goes (RTG), and predicts the next action. Despite the appealing architectural innovation, subsequent studies have exposed two major issues with DT: (1) failure in the face of high environmental stochasticity \citep{Emmons2021Rvs, paster2022CountOnLuck} and (2) the inability to stitch suboptimal trajectories \citep{yamagata2023QLearningDT, hu2024QRegularizedTransformer}.

Recently, \citet{Bhargava2024WhenDT} provided a thorough comparison of DT with conventional MLP-based offline RL algorithms, including Behavior Cloning (BC) and Conservative Q-Learning (CQL) \citep{Kumar2020CQL}. They primarily examined two offline datasets: the D4RL \citep{fu2020d4rl} dataset and the Robomimic \citep{robosuiteDocs} dataset. For both datasets, they consider both dense reward and sparse reward variations. They conclude from their experiments that DT is a preferable algorithm in sparse-reward, low-quality data, and long-horizon settings, when compared with vanilla Behavior Cloning (BC) and Conservative Q-learning (CQL) \citep{Kumar2020CQL}.

\section{Problem Setup}
In this paper, we aim to explore whether DT is preferable in sparse reward environments. Following the experimental design of \citet{Bhargava2024WhenDT}, we consider two sparse-reward settings. 

\paragraph{Sparse Reward Setting}  
This setting corresponds to environments where binary rewards are assigned only at the terminal timestep of trajectories. Let $\mathcal{D} = \{\tau_i\}_{i=1}^N$ denote a dataset of $N$ trajectories, where each trajectory $\tau_i = \{(s_t^i, a_t^i, r_t^i)\}_{t=1}^{T_i}$ consists of states, actions, and rewards. The reward function is defined as:
\[
r_t^i =
\begin{cases}
1, & \text{if } t = T_i \text{ and } \tau_i \text{ is successful}, \\
0, & \text{otherwise}.
\end{cases}
\]
For example, in a dataset with $N = 300$ trajectories, if 100 are labeled successful, then each of these 100 trajectories receives a cumulative reward of 1; the remaining 200 receive 0.
% \paragraph{Sparsified Reward Setting.} This setting is derived from environments that originally provide dense per-timestep rewards. In offline reinforcement learning, where agents train on static datasets without further interaction with the task environment, sparse-reward conditions can be simulated by post-processing the dataset.
% Specifically, following \citet{Bhargava2024WhenDT}, we set 
% all intermediate rewards to zero and retain only the terminal reward in each trajectory. Specifically, a trajectory with rewards \( r_0, r_1, \ldots, r_{T-1} \) is modified such that
% \( r_t \leftarrow 0 \) for all \( t \leq T-1 \), and \( r_{T-1} \leftarrow r_{T-1}\). This allows the offline learning agent to experience a reward distribution analogous to that of truly sparse environments. The key distinction is that in sparse settings, terminal rewards are inherently binary (typically 0 or 1), whereas in sparsified settings, terminal rewards retain their original non-binary, task-specific values.

% % \dzx{So we only retain successful trajs?}\ym{No. suppose in a dataset trajectories have rewards r0, r1, ..., r_t, from t=0 to t=t-1, let the rewards all be 0, and at the end of the trajectory it has non-binary number at the terminal of the trajectory}

\paragraph{Sparsified Reward Setting} This setting is derived from environments that originally provide dense per-timestep rewards. In offline reinforcement learning, where agents train on static datasets without further interaction with the task environment, sparse-reward conditions can be simulated by post-processing the dataset. Specifically, following \citet{Bhargava2024WhenDT}, we set all intermediate rewards to zero and move the total return of each trajectory to the final timestep. That is, for a trajectory with rewards \( r_0, r_1, \ldots, r_{T-1} \), we modify it such that \( r_t \leftarrow 0 \) for all \( t < T-1 \), and \( r_{T-1} \leftarrow \sum_{t=0}^{T-1} r_t \). This allows the offline learning agent to experience a reward distribution analogous to that of truly sparse environments. The key distinction is that in sparse settings, terminal rewards are inherently binary (typically 0 or 1), whereas in sparsified settings, terminal rewards retain their original non-binary, task-specific values.

\section{Methods}
\label{sec:algorithms}

\citet{Bhargava2024WhenDT} compares Behavior Cloning (BC), Conservative Q-Learning (CQL) \citep{Kumar2020CQL}, and Decision Transformer (DT) \citep{Chen2021DecisionTransformer}. BC and CQL run over an MLP backbone, whereas DT runs over a transformer backbone. We also introduce two new methods: Filtered Behavior Cloning (FBC), which runs over MLP backbones, and Filtered Decision Transformer (FDT), which runs over transformer backbones. 

% \ym{I change the above to the following, let me know what you think}

% \citet{Bhargava2024WhenDT} compare three offline-RL baselines: Behavioural Cloning (BC), Conservative Q-Learning (CQL; \citep{Kumar2020CQL}), and Decision Transformer (DT; \citep{Chen2021DecisionTransformer}). In their implementation BC and CQL share the same multilayer-perceptron (MLP) encoder, whereas DT replaces this encoder with a causal transformer while retaining the original MLP prediction heads. We further propose two variants that apply a temporal-filtering step: Filtered Behavioural Cloning (FBC), which keeps the MLP backbone, and Filtered Decision Transformer (FDT), which keeps the transformer backbone. 

% Because BC and CQL are well established, we do not restate their derivations here.

\paragraph{Decision Transformer (DT)} During inference, DT chooses the next action based on a fixed-length sequence of target return-to-go, state, and action triplets from prior timesteps. Given a context length $K$, the action prediction depends on the temporally ordered sequence $\{(R_j, s_j, a_j)\}_{j=t-K+1}^{t}$, where $R_j$ is the target return-to-go. The model processes this sequence through a transformer encoder that uses sinusoidal positional encodings to preserve temporal order, and stacked self-attention layers to model dependencies. One of the critical issues in DT is setting the target return to go. In the sparse-reward setting, setting the target to one is a natural choice. However, in the sparsified-reward setting, it is less obvious. 

\paragraph{Filtered Behavior Cloning (FBC)} 
We now consider a very simple algorithm, Filtered Behavior Cloning (FBC). In FBC, we first process the offline dataset by retaining only the high-performing trajectories. \emph{Then we simply apply vanilla BC to the resulting dataset.} When we refer to FBC, we assume that the underlying backbone is a Multi-Layer Perceptron (MLP). 

The filtering rule is as follows. For the sparse-reward setting, we only retain the trajectories that are successful. For the sparsified-reward setting, we only retain trajectories for which the final return $r_{T-1}$ is in the top $x\%$ of all the final returns in the dataset. In this paper, we use $x = 10\%$.

\paragraph{Filtered Decision Transformer (FDT).} Additionally, we also consider training DT only on the high-performing trajectories, where we filter the trajectories exactly the same way we filter the trajectories for FBC. Although the common belief is that DT learns from both successful and unsuccessful trajectories, including this variant allows us to test whether DT truly benefits from this design in sparse-reward scenarios.

\section{Experiments}
% In this work, we aim to evaluate task performance in sparse robotic manipulation settings. Building on the foundation established in the original study, we investigate the sparse tasks in ROBOMIMIC, specifically ``PickPlaceCan'' and ``Lift,'' within the Robosuite simulation environment. These tasks utilize proficient (PH), multi-human (MH), and machine-generated (MG) datasets, where behavior cloning (BC) demonstrated superior performance on PH and MH datasets, and decision transformers (DT) excelled on MG datasets. Additionally, we replicate experiments on sparsified D4RL locomotion tasks, including Walker2d, Hopper, and Half-Cheetah, which feature medium, medium-expert, and medium-replay datasets. Sparse modifications involve removing intermediate rewards and providing only a cumulative numerical reward at the final step.

% We evaluate all algorithms introduced in Section~\ref{sec:algorithms} under two sparse reward settings: (1) sparsified locomotion tasks from D4RL, and (2) Robomimic tasks in the robosuite environment. \dzx{Citations} Our experimental design follows \citet{Bhargava2024WhenDT} for comparability. All results \dzx{What metric? What are the bold numbers} are averaged over 5 seeds, and evaluation is conducted every 50 epochs using 50 rollouts.

We evaluate all algorithms introduced in Section~\ref{sec:algorithms} for two classes of sparse datasets : (1) a sparsified version of the D4RL dataset \citep{fu2020d4rl}, and (2) the sparse versions of the Robomimic tasks in the Robosuite environment~\citep{mandlekar2021robomimic, zhu2020robosuite}. Our experimental design closely follows \citet{Bhargava2024WhenDT} for comparability. Results reported for DT, CQL, and BC were obtained with exactly the same parameters reported in \citet{Bhargava2024WhenDT}.  All results are averaged over 5 seeds, and evaluation is conducted every 50 epochs using 50 rollouts. 
%Reported scores reflect the best (i.e., maximum) success rate achieved during training, consistent with the evaluation protocol in \citet{Bhargava2024WhenDT}. 
Bolded values indicate the highest-performing method for each task.

\subsection{Sparsified D4RL Locomotion}
\label{sec:d4rl-results}
For the D4RL sparsified setting, we evaluate on the locomotion tasks: walker2d, hopper, and halfcheetah, using sparsified versions of the medium, medium-replay, and medium-expert datasets. Further details are included in Appendix~\ref{appendix:d4rl}.

\begin{table*}[htbp]
  \caption{Comparison of BC, CQL, DT, FBC, and FDT across different D4RL environments. Results show the average normalized D4RL score and its standard deviation.}
  \small
  \centering
  \scalebox{0.8}{
  \begin{tabular}{cccccc}
    \toprule
    \textbf{Dataset} & \textbf{BC} & \textbf{CQL} & \textbf{DT} & \textbf{FBC} & \textbf{FDT} \\
    \midrule
    Half Cheetah - Medium & $42.76\pm0.17$ & $38.63 \pm 0.81$ & $42.65 \pm 1.05$ & \textbf{43.51} $\pm$ \textbf{1.35} & $42.17 \pm 1.96$ \\
    Hopper - Medium        & $64.35 \pm 5.6$ & \textbf{73.89} $\pm$ \textbf{10.12} & 73.40 $\pm$ 11.92 & $61.12 \pm 10.13$ & $63.23 \pm 12.28$ \\
    Walker - Medium        & $54.62\pm12.04$ & $19.31 \pm 3.17$ & $73.28 \pm 11.65$ & \textbf{77.66} $\pm$ \textbf{9.61} & $75.24 \pm 13.60$ \\
    \midrule
    \textbf{Medium Average} & 53.91 $\pm$ 5.93  & $43.94 \pm 4.70$ & \textbf{63.11} $\pm$ \textbf{8.21} & $60.76 \pm 7.03$ & $60.21 \pm 9.28$ \\
    \midrule
    Half Cheetah - Medium Replay & $9.81\pm9.2$ & $35.00 \pm 2.56$ & $39.45 \pm 2.43$ & \textbf{41.83} $\pm$ \textbf{2.44} & $33.70 \pm 7.79$ \\
    Hopper - Medium Replay      & $16.19 \pm 10.8$ & 83.10 $\pm$ 19.21 & $71.59 \pm 10.18$ & \textbf{93.53} $\pm$ \textbf{8.05} & $82.74 \pm 13.33$ \\
    Walker - Medium Replay      & $17.82\pm4.96$ & $29.02 \pm 19.63$ & $65.20 \pm 14.31$ & \textbf{72.69} $\pm$ \textbf{19.54} & $65.10 \pm 17.58$ \\
    \midrule
    \textbf{Medium-Replay Average} & 14.6 $\pm$ 8.32 & $49.04 \pm 13.79$ & $58.75 \pm 8.97$ & \textbf{69.35} $\pm$ \textbf{10.01} & $60.51 \pm 12.90$ \\
    \midrule
    Half Cheetah - Medium Expert & $42.95\pm0.14$ & $24.35 \pm 2.38$ & \textbf{93.66} $\pm$ \textbf{1.01} & $92.99 \pm 0.98$ & $78.24 \pm 20.39$ \\
    Hopper - Medium Expert      & $62.21 \pm 6.5$  & $42.44 \pm 12.52$ & $111.37 \pm 1.02$ & \textbf{111.46} $\pm$ \textbf{0.83} & $106.52 \pm 13.28$ \\
    Walker - Medium Expert      & $38\pm9.86$ & $21.30 \pm 0.55$ & $107.90 \pm 0.98$ & \textbf{109.21} $\pm$ \textbf{0.29} & \textbf{109.21} $\pm$ \textbf{0.51} \\
    \midrule
    \textbf{Medium-Expert Average} & 47.72 $\pm$  5.5 & $29.36 \pm 5.14$ & $104.31 \pm 1.00$ & \textbf{104.55} $\pm$ \textbf{0.70} & $97.99 \pm 11.39$ \\
    \midrule
    \textbf{Total Average} & 38.74 $\pm$ 6.58 & $40.78 \pm 7.88$ & $75.39 \pm 6.06$ & \textbf{78.22} $\pm$ \textbf{5.91} & $72.90 \pm 11.19$ \\
    \bottomrule
  \end{tabular}
  }
  \label{tab:d4rl-results}
\end{table*}

Table~\ref{tab:d4rl-results} shows the final performance of the various algorithms. As also reported in \citet{Bhargava2024WhenDT}, we see DT does significantly better than both BC and CQL, confirming that DT is indeed preferable to CQL for the sparsified-reward version of D4RL. \emph{However, we also see that FBC has a higher total average than DT, and that FBC beats DT in 7 of the 9 datasets.} 
%Therefore, at least for D4RL, the very simple FBC algorithm with MLP backbone is preferable to the more complex DT algorithm with a transformer backbone. 
Interestingly, FDT does a little worse than DT (and hence worse than FBC), which seems to indicate that DT can learn from both high-quality and low-quality trajectories. 

%CQL is included for baseline reference but is not discussed further. FBC and DT perform comparably overall, with FBC leading in all medium-replay settings. FDT shows less consistent performance.

\subsection{Robomimic Sparse Setting}

\label{sec:robomimic-results}
In the Robomimic sparse setting, as in \citet{Bhargava2024WhenDT}, we use the machine-generated (MG) datasets from the Robomimic benchmark built on Robosuite. These datasets are characterized by low-quality demonstrations. Following \citet{Bhargava2024WhenDT}, we consider two tasks: Lift and PickPlaceCan. Details on the tasks and dataset configuration are provided in Appendix~\ref{appendix:robomimic}.

\begin{table}[htbp]
    \caption{Comparison of BC, CQL, DT, FBC, and FDT methods on the Lift and Can tasks. The best success rate is reported.}
    \centering
    \scalebox{0.8}{
    \begin{tabular}{lccccc}
    \toprule
    Dataset & BC & CQL & DT & FBC & FDT \\
    \midrule
    Lift    & 0.46 $\pm$ 0.05 & 0.60 $\pm$ 0.13  & 0.92 $\pm$ 0.03 & \textbf{0.94 $\pm$ 0.03} & 0.93 $\pm$ 0.05 \\
    Can     & 0.45 $\pm$ 0.09 & $0\pm0$ & 0.79 $\pm$ 0.06 & 0.84 $\pm$ 0.04 & \textbf{0.89 $\pm$ 0.03} \\
    \midrule
    \textbf{Average} & 0.45 $\pm$ 0.07 & 0.30 $\pm$ 0.07 & 0.86 $\pm$ 0.05 & 0.89 $\pm$ 0.04 & \textbf{0.91 $\pm$ 0.04} \\
    \bottomrule
    \end{tabular}
    }
    \label{tab:comparison_best_score}
\end{table}

Table~\ref{tab:comparison_best_score} summarizes the best evaluation achieved during training. As also reported in \citet{Bhargava2024WhenDT}, we see DT does significantly better than both BC and CQL, confirming that DT is indeed preferable to CQL for these sparse-reward datasets. \emph{However, we also see that DT, FBC, and FDT all have similar performance, with FBC being a little higher than DT on both datasets, and with FBC also higher than FDT on Lift and lower on Can.} 
 
\section{Discussion}

We have just established that for the sparse datasets considered in \citet{Bhargava2024WhenDT}, FBC has better performance than DT. We also observed that the training wall-clock time for DT is about three times longer than it is for FBC. Also, the transformer employed for DT has approximately one million parameters, whereas the MLP used for FBC has about half that number. \emph{For sparse reward environments, we therefore conclude that DT is {\bf not} preferable.}

Why is it that DT does not beat simple FBC? As a thought experiment, suppose that DT did not have the returns-to-go in training or inference. And suppose that the state is Markovian, or can be made nearly Markovian by defining the state as the most recent $k$ states and actions. In this case, we would expect the DT policy to resemble the BC policy obtained with an MLP. In the sparse-reward setting, by including the return-to-go in DT, we are doing no more than indicating to DT which training trajectories are good and which are bad. Thus, intuitively, DT in the sparse reward setting would generate policies that are similar to FBC. In our experiments, we show that DT actually performs somewhat worse than FBC. This could be due to a number of factors, including overfitting and poor credit assignment. 

If DT is not preferable for sparse-reward environments, is it preferable for dense-reward? \citep{Bhargava2024WhenDT} shows that CQL beats DT for the original (i.e., not sparsified) D4RL datasets. And although we do not provide empirical evidence here, prior literature implies that DT is also not preferable for diverse dense-reward datasets in general \citep{Emmons2021Rvs, tarasov2023corl, yamagata2023QLearningDT, hu2024QRegularizedTransformer, Bhargava2024WhenDT}. Thus, we argue that \emph{DT is barely preferable for raw-state robotic tasks in offline RL}.

\section{Conclusion}

The potential contributions of Decision Transformer (DT) can be broadly categorized into two aspects: first, conditioning on return-to-go (RTG); and second, modeling long-range temporal dependencies via attention mechanisms \citep{Chen2021DecisionTransformer}. However, increased algorithmic and architectural sophistication does not inherently lead to improved performance in offline reinforcement learning. In sparse-reward domains, where learning signals are limited by nature, the inductive biases built into generic transformer architectures like DT often fail to yield concrete advantages over the basic MLP trained on selectively filtered trajectories. 

% Learning from successful demonstrations has long been a core focus in imitation learning \citep{Ho2016GAIL, Hussein2017ImitationSurvey}. Methods such as Best-Action Imitation Learning (BAIL) \citep{Chen2020BAIL} and Value-Aligned Behavior Cloning \citep{Jiang2024ValueAligned} explicitly exploit high-quality trajectories to guide policy learning. In this vein, our implementation of Filtered Behavior Cloning (FBC), a simple multilayer perceptron trained exclusively on successful episodes, shares conceptual foundations with this class of approaches.

Our critique is not intended as a wholesale dismissal of transformer-based models. When carefully tailored to the structure of sequential decision-making, transformers can offer genuine benefits. For example, the Behavior Transformer \citep{Shafiullah2022BehaviorTransformer} incorporates architectural enhancements to better model multimodal behaviors. Similarly, the Graph Decision Transformer (GDT) \citep{hu2023graphdt} recasts trajectories as causal graphs, mitigating reliance on potentially uninformative RTG signals and achieving strong results on vision-based tasks such as Atari and Gym. More recently, the Q-value Regularized Decision Transformer (QT) \citep{hu2024QRegularizedTransformer} fuses dynamic programming insights with sequence modeling, consistently outperforming both standard DTs and classical DP-based methods on D4RL benchmarks.

Our empirical analysis, together with \citet{Bhargava2024WhenDT}, calls for ongoing and continuing algorithmic and architectural improvements on top of the vanilla DT to fully unleash the power of the transformer architecture for both dense-reward and sparse-reward RL in future studies.

\newpage

%%%%%%%%%%%%%%%%%%%%%%%%%%%%%%%%%%%%%%%%%%%%%%%%%%%%%%%%%%%%%%%%
%% Appendices
%%%%%%%%%%%%%%%%%%%%%%%%%%%%%%%%%%%%%%%%%%%%%%%%%%%%%%%%%%%%%%%%
\appendix

\section{Experiment Details}
\label{sec:appendix1}
\subsection{Robomimic and Robosuite}
\label{appendix:robomimic}

% this figure should be put to the ablation studies about the effect on BC.
\begin{figure}[htbp]
    \centering
    \scalebox{1.0}{
    \begin{subfigure}[b]{0.45\textwidth}
        \centering
        \includegraphics[width=\textwidth]{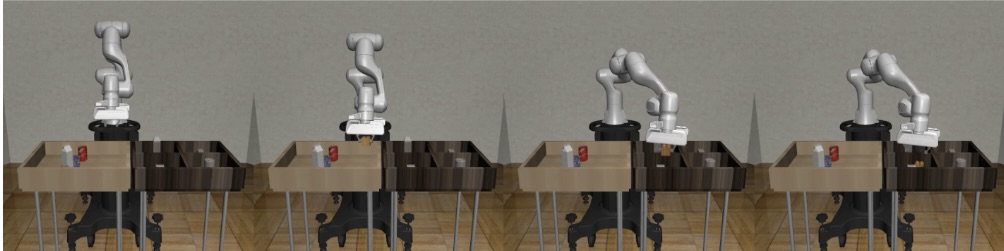}
        \caption{PickPlaceCan}
        \label{fig:PickPlaceCan}
    \end{subfigure}
    \hfill
    \begin{subfigure}[b]{0.45\textwidth}
        \centering
        \includegraphics[width=\textwidth]{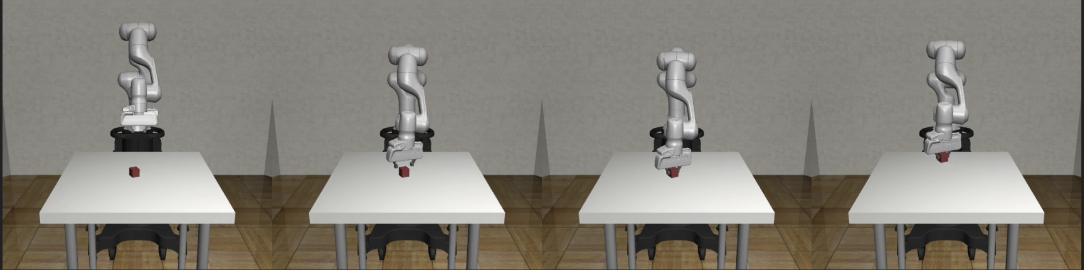}
        \caption{Lift}
        \label{fig:Lift}
    \end{subfigure}
    }
    \caption{Illustrations of the \textit{Lift} and \textit{PickPlaceCan} tasks in Robosuite \citep{robosuiteDocs}.}
    \label{fig:appendix_robomimic}
\end{figure}

We evaluate algorithms in sparse-reward robotic manipulation tasks using the \textbf{Robosuite} simulator \citep{zhu2020robosuite} and the \textbf{Robomimic} dataset suite \citep{mandlekar2021robomimic}. Robosuite is a modular simulation environment supporting both dense and sparse reward configurations. Following the experimental setup of \citet{Bhargava2024WhenDT}, who examined both reward regimes on the \textit{Lift} and \textit{PickPlaceCan} tasks, we focus exclusively on the sparse setting to align with the central objectives of our study. Visual illustrations of the two tasks are provided in Figure~\ref{fig:appendix_robomimic}, and task details are summarized in Table~\ref{tab:robosuite-task-summary-transposed}. 

Robomimic is a standardized benchmark offering human- and machine-generated demonstration datasets. It includes three variants: Proficient Human (PH), Multi-Human (MH), and Machine-Generated (MG). As shown in \citet{Bhargava2024WhenDT}, behavior cloning performs well on PH and MH datasets, while Decision Transformer shows stronger results on the MG variant. Based on these findings, we restrict our experiments to the MG dataset, which contains the lowest-quality demonstrations. This dataset is constructed by sampling rollouts from Soft Actor-Critic (SAC) agents at multiple training checkpoints, thus offering a diverse but suboptimal data distribution.

\begin{table}[ht]
\centering
\caption{Robosuite Tasks and Dataset Specifications}
\label{tab:robosuite-task-summary-transposed}
\scalebox{0.85}{
\begin{tabular}{p{4cm}p{5.5cm}p{5.5cm}}
\toprule
\textbf{Attribute} & \textbf{Lift} & \textbf{PickPlaceCan} \\
\midrule
\textbf{Scene Description} 
& A cube is placed on a table in front of a robot arm. The cube’s position is randomized each episode. 
& A bin with four objects is placed in front of the robot, with four nearby target containers. Object positions are randomized per episode. \\
\midrule
\textbf{Objective} 
& Lift the cube vertically above a predefined height. 
& Pick each object and place it into its corresponding container. \\
\midrule
\textbf{Dataset Composition} 
& 244 / 1500 trajectories (300 rollouts × 5 checkpoints), machine-generated via SAC. 
& 716 / 3900 trajectories (300 rollouts × 13 checkpoints), machine-generated via SAC. \\
\midrule
\textbf{Observation \& Action Space} 
& Observations: 18D proprioceptive input \newline
Actions: 7D Cartesian EE + gripper control 
& Observations: 22D proprioceptive input \newline
Actions: 7D Cartesian EE + gripper control \\
\bottomrule
\end{tabular}
}
\end{table}

\subsection{D4RL Environments and Datasets}
\label{appendix:d4rl}

We perform evaluations on continuous control tasks from the \textbf{D4RL benchmark suite} \citep{fu2020d4rl}, a widely adopted standard for offline reinforcement learning. Accordingly, we omit detailed elaboration here. Our experiments focus on three locomotion tasks, \textit{Hopper}, \textit{Walker2d}, and \textit{HalfCheetah}, using the \textit{Medium}, \textit{Medium-Replay}, and \textit{Medium-Expert} dataset variants, following the setup in \citet{Bhargava2024WhenDT}. 
% Learning curves comparing DT, FDT, and FBC across the nine task-dataset combinations are presented in Figure~\ref{fig:mujoco_sub}. CQL performance curves are excluded since (1) \citet{Bhargava2024WhenDT} already provides a comprehensive performance evaluation of CQL; and (2) it diverts from the core comparative focus of our study.

\subsection{Hyperparameters}
\label{appendix:hyperparameters}

We adopt the hyperparameter setup as \citet{Bhargava2024WhenDT} and report everything in Table~\ref{tab:dt_bc_fbc_cql_hyperparams} for reproducibility.

\begin{table}[ht]
    \caption{Hyperparameters for DT, BC, FBC, and CQL evaluations.}
    \label{tab:dt_bc_fbc_cql_hyperparams}
    \centering
    \scalebox{0.7}{
    \begin{tabular}{lll}
    \toprule
    \textbf{Category} & \textbf{Hyperparameter} & \textbf{Value} \\
    \midrule
    \multirow{8}{*}{\textbf{Transformer (DT)}} 
        & Number of layers & 3 \\
        & Attention heads & 1 \\
        & Embedding dimension & 128 \\
        & Nonlinearity & ReLU \\
        & Context length & 1 (Robomimic), 20 (D4RL) \\
        & Dropout & 0.1 \\
        & Return-to-go (RTG) conditioning & 120 (Robomimic), 6000 (HalfCheetah), 3600 (Hopper), 5000 (Walker) \\
        & Max episode length & 1000 \\
    \midrule
    \multirow{3}{*}{\textbf{MLP (BC)}} 
        & Network depth & 2 layers \\
        & Hidden units per layer & 512 \\
        & Nonlinearity & ReLU \\
    \midrule
    \multirow{7}{*}{\textbf{Training (DT, BC)}} 
        & Batch size & 512 (DT), 100 (BC) \\
        & Learning rate & $10^{-4}$ \\
        & Learning rate decay & 0.1 (BC) \\
        & Grad norm clip & 0.25 \\
        & Weight decay & $10^{-4}$ \\
        & LR scheduler & Linear warmup for first $10^5$ steps (DT) \\
        & Epochs & 100 \\
    \midrule
    \multirow{5}{*}{\textbf{Evaluation}} 
        & Frequency & Every 50 epochs (Robomimic), Every 100 epochs (D4RL) \\
        & Rollouts per eval & 50 \\
        & Evaluation episodes & 100 \\
        & Seeds & 5 \\
        & Reference (DT) & \url{https://github.com/kzl/decision-transformer} \\
    \midrule
    \multirow{17}{*}{\textbf{CQL}} 
        & Batch size & 2048 \\
        & Steps per iteration & 1250 \\
        & Iterations & 100 \\
        & Discount factor & 0.99 \\
        & Policy learning rate & $3 \times 10^{-4}$ \\
        & Q-function learning rate & $3 \times 10^{-4}$ (D4RL), $1 \times 10^{-3}$ (Robomimic) \\
        & Actor MLP dimensions & [300, 400] (Robomimic) \\
        & Soft target update rate & $5 \times 10^{-3}$ \\
        & Target update period & 1 \\
        & Alpha multiplier & 1 \\
        & CQL n\_actions & 10 \\
        & Min Q weight & 5 \\
        & CQL temperature & 1 \\
        & Importance sampling & True \\
        & Lagrange tuning & False \\
        & Target action gap & -1 \\
        & Lagrange threshold $\tau$ & 5 (Robomimic) \\
        & Reference (CQL) & \url{https://github.com/tinkoff-ai/CORL} \\
    \bottomrule
    \end{tabular}
    }
\end{table}

\clearpage
\section{Learning Curves}
\label{appendix:learning-curves}

\begin{figure}[htbp]
    \centering
    \scalebox{0.8}{
    \begin{subfigure}[b]{0.45\textwidth}
        \centering
        \includegraphics[width=\textwidth]{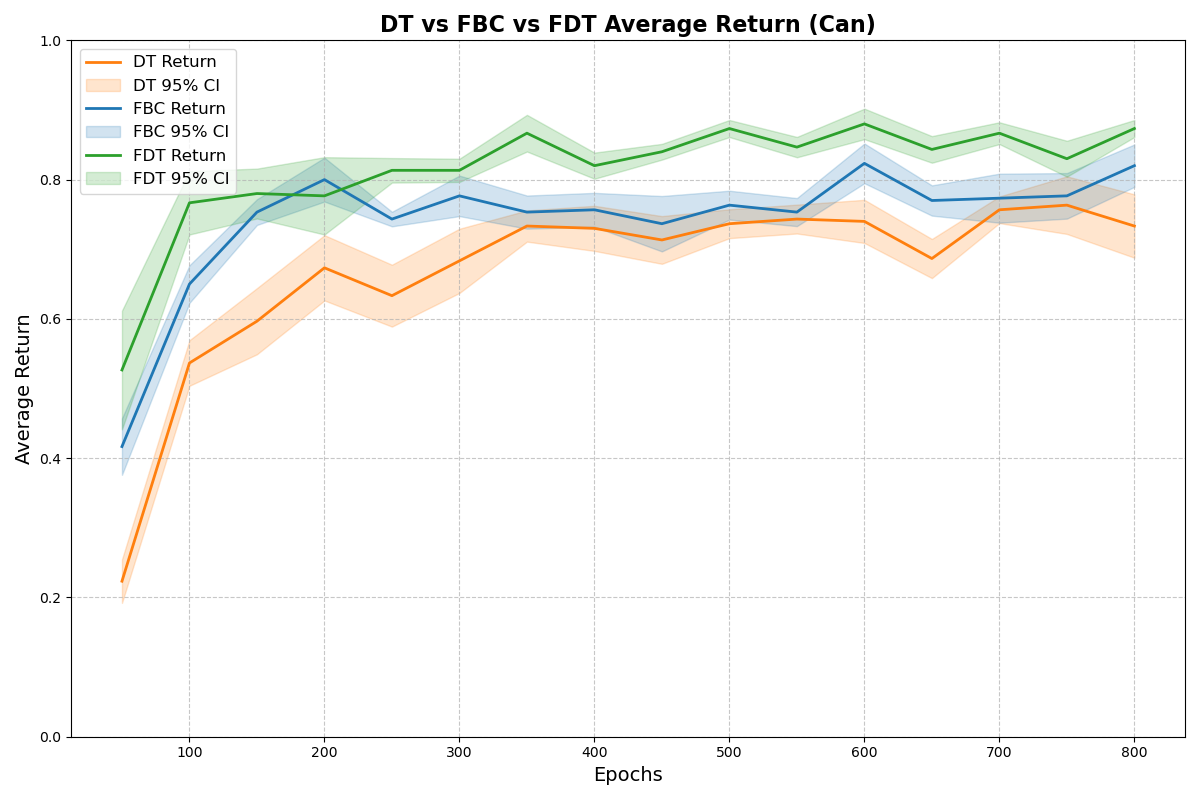}
        \caption{Results for the task PickupCan}
        \label{fig:can_filtered_comparison}
    \end{subfigure}
    \hfill
    \begin{subfigure}[b]{0.45\textwidth}
        \centering
        \includegraphics[width=\textwidth]{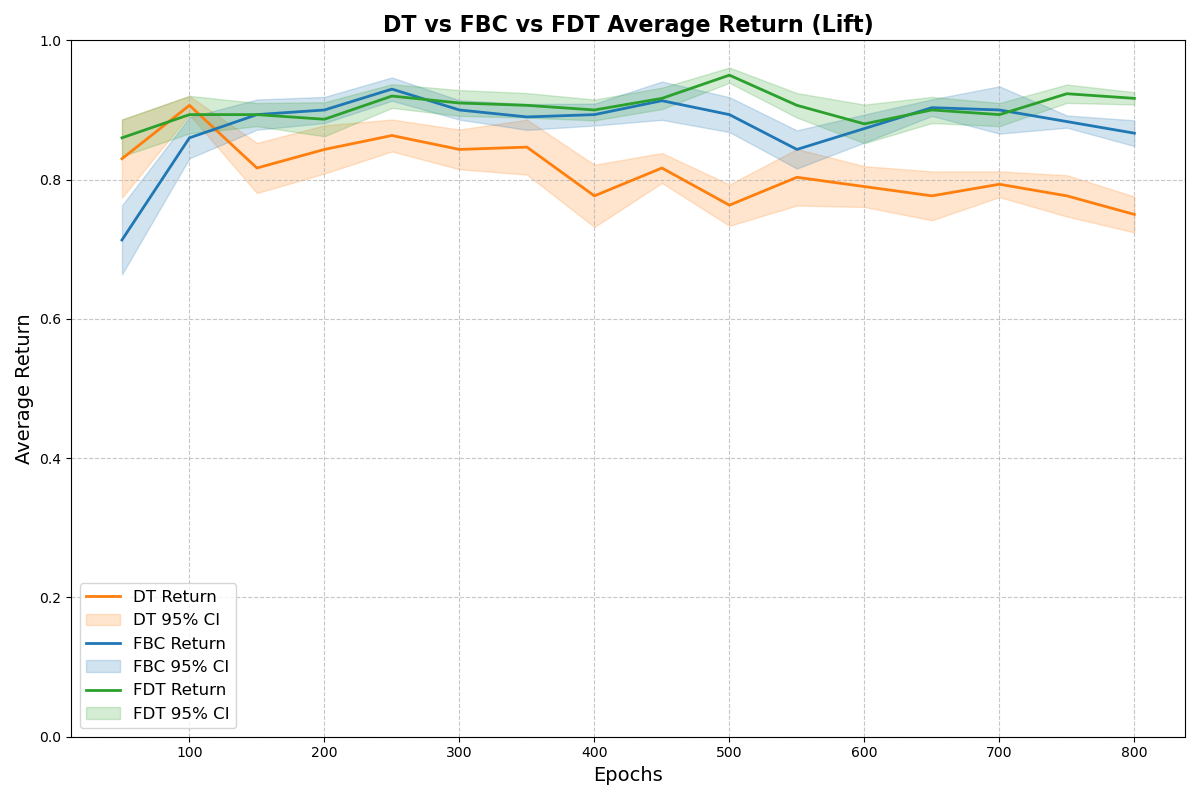}
        \caption{Results for the task Lift}
        \label{fig:lift_filtered_comparison}
    \end{subfigure}
    }
    \caption{Results for Robomimic tasks. Performance comparison of FBC, FDT, DT.}
    \label{fig:robomimic_main}
\end{figure}

\begin{figure}[ht]
    \centering

    \begin{subfigure}[b]{0.32\textwidth}
        \centering
        \includegraphics[width=\textwidth]{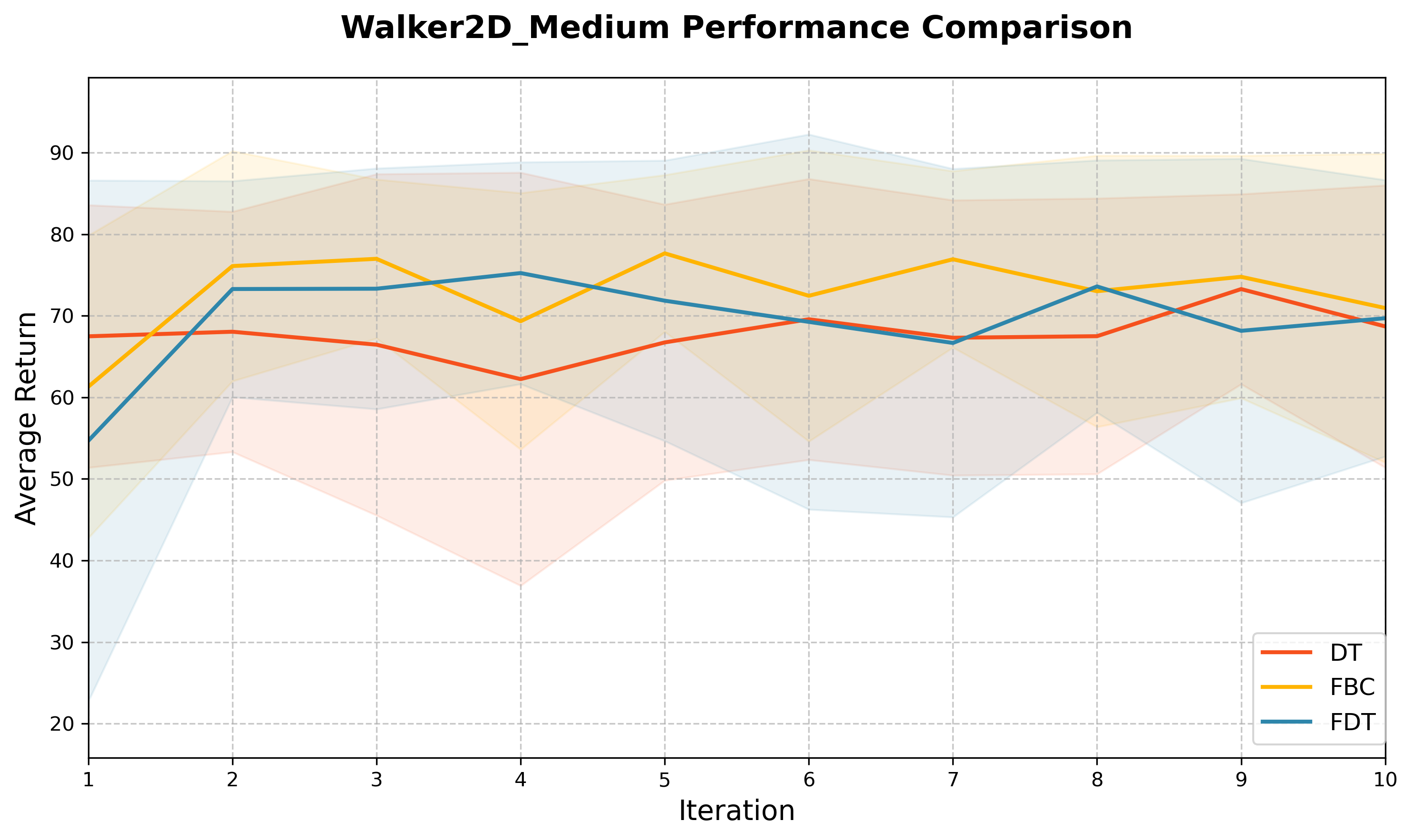}
        \caption{Walker2d (Medium)}
        \label{fig:walker2d_medium_comparison}
    \end{subfigure}
    \hfill
    \begin{subfigure}[b]{0.32\textwidth}
        \centering
        \includegraphics[width=\textwidth]{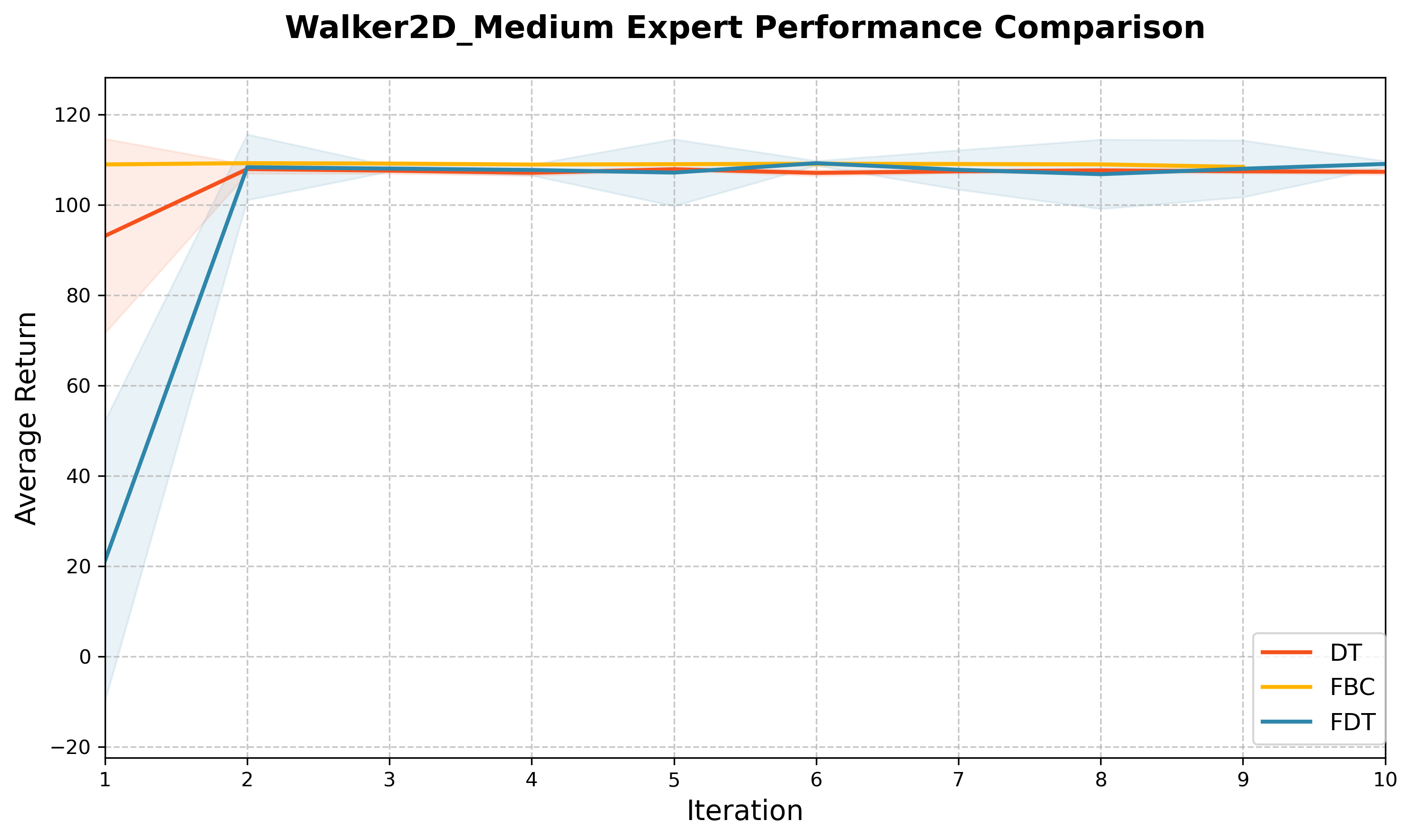}
        \caption{Walker2d (Medium-Expert)}
        \label{fig:walker2d_medium_expert_comparison}
    \end{subfigure}
    \hfill
    \begin{subfigure}[b]{0.32\textwidth}
        \centering
        \includegraphics[width=\textwidth]{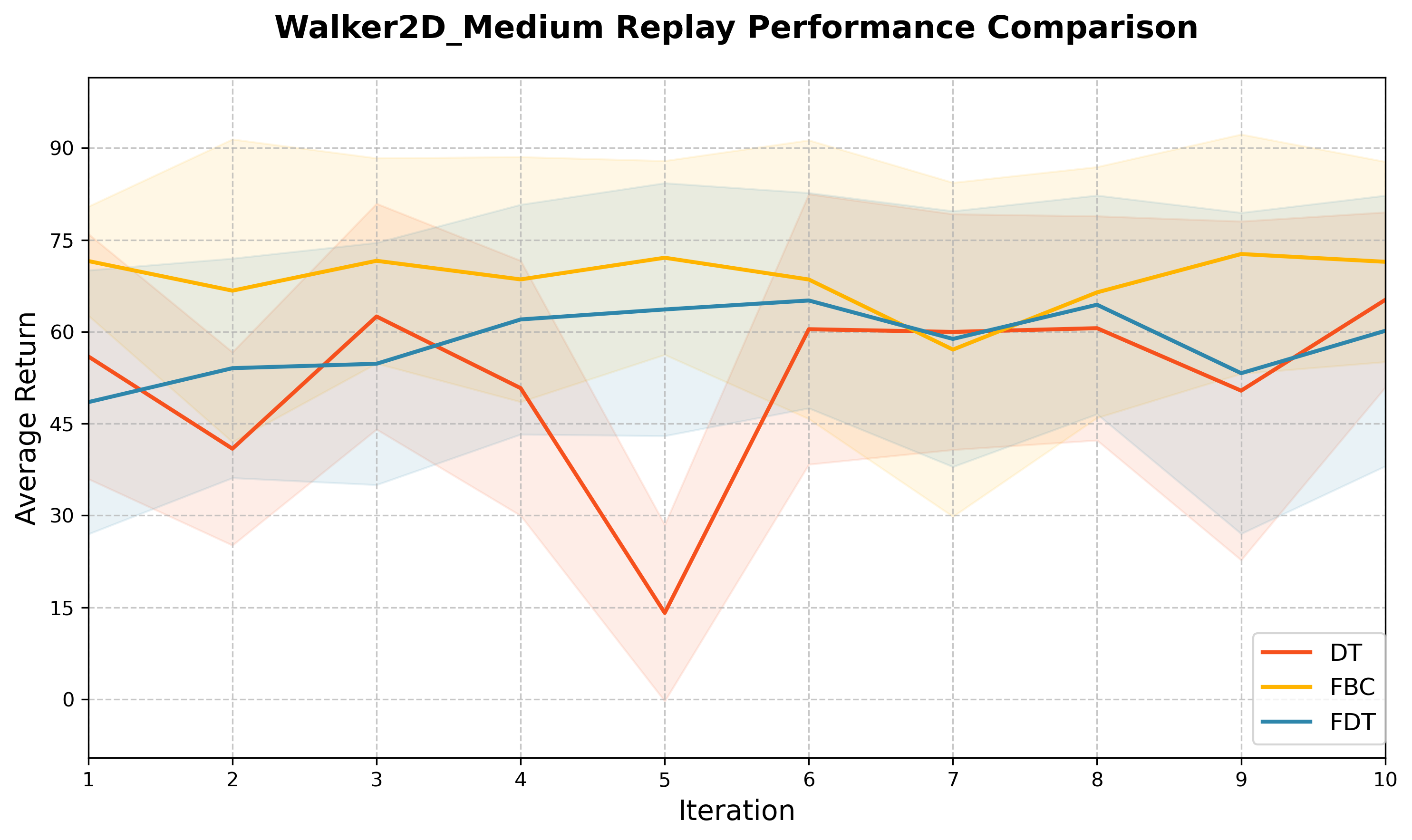}
        \caption{Walker2d (Medium-Replay)}
        \label{fig:walker2d_medium_replay_comparison}
    \end{subfigure}
    \par
    \vspace{0.5cm}
    \begin{subfigure}[b]{0.32\textwidth}
        \centering
        \includegraphics[width=\textwidth]{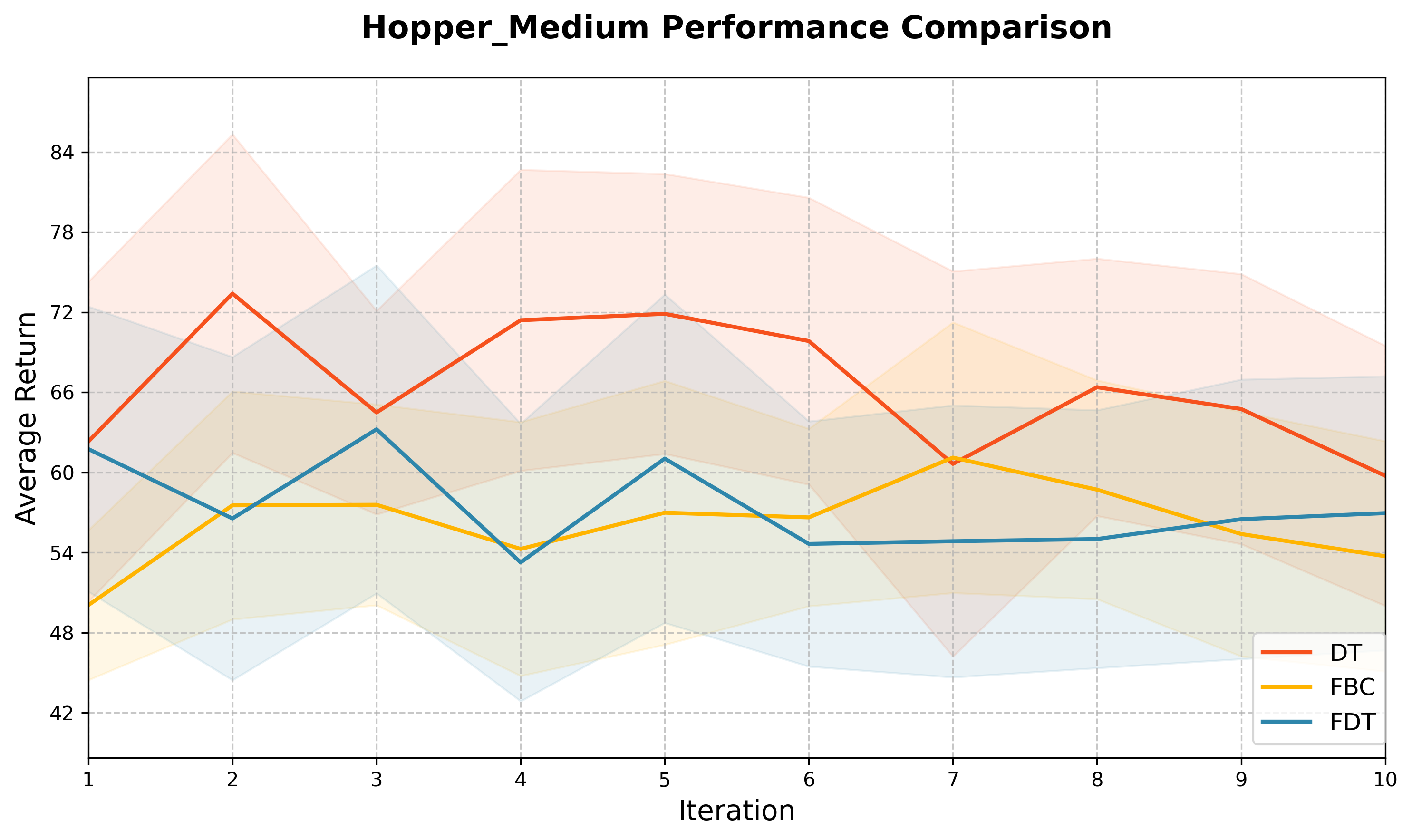}
        \caption{Hopper (Medium)}
        \label{fig:hopper_medium_comparison}
    \end{subfigure}
    \hfill
    \begin{subfigure}[b]{0.32\textwidth}
        \centering
        \includegraphics[width=\textwidth]{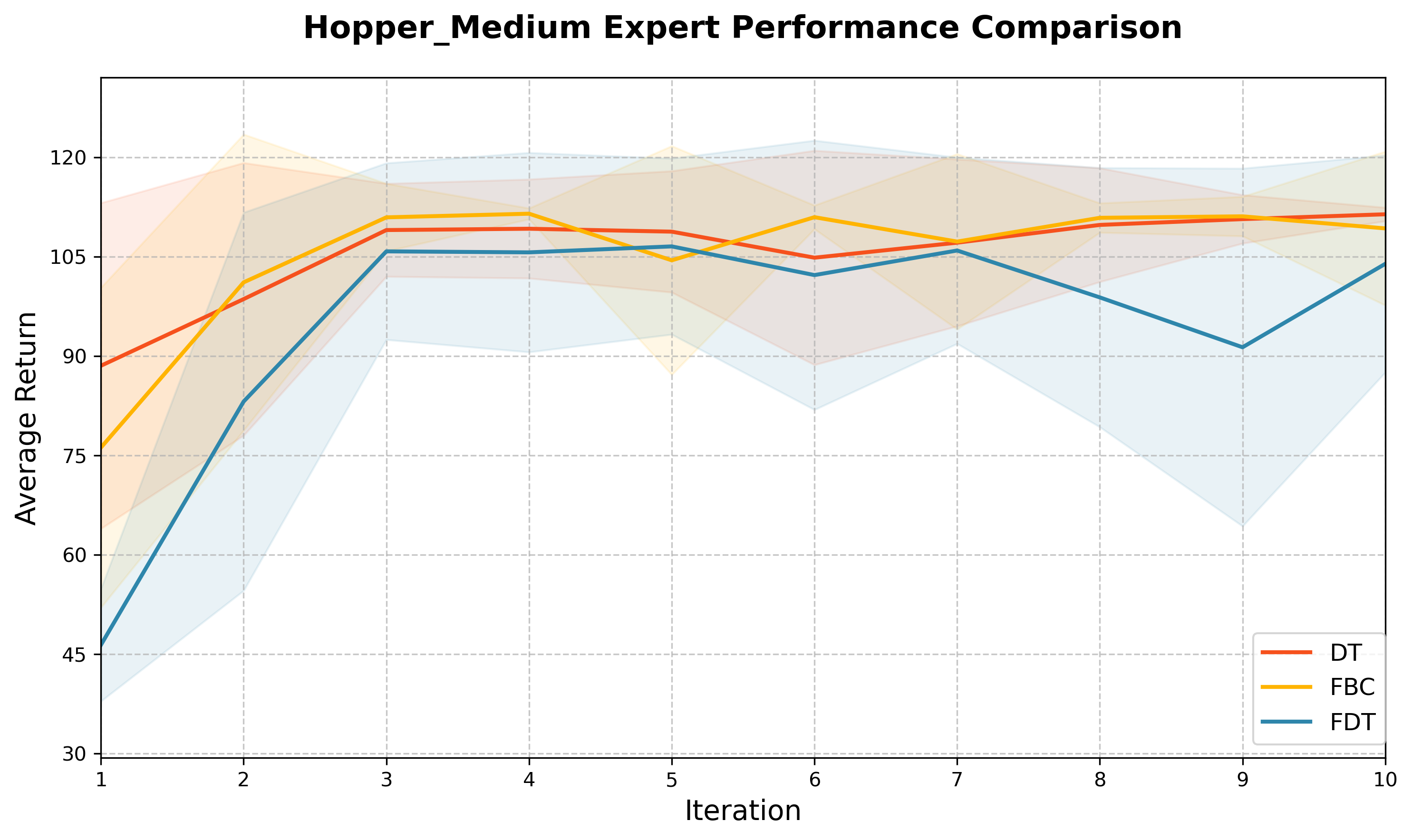}
        \caption{Hopper (Medium-Expert)}
        \label{fig:hopper_medium-expert_comparison}
    \end{subfigure}
    \hfill
    \begin{subfigure}[b]{0.32\textwidth}
        \centering
        \includegraphics[width=\textwidth]{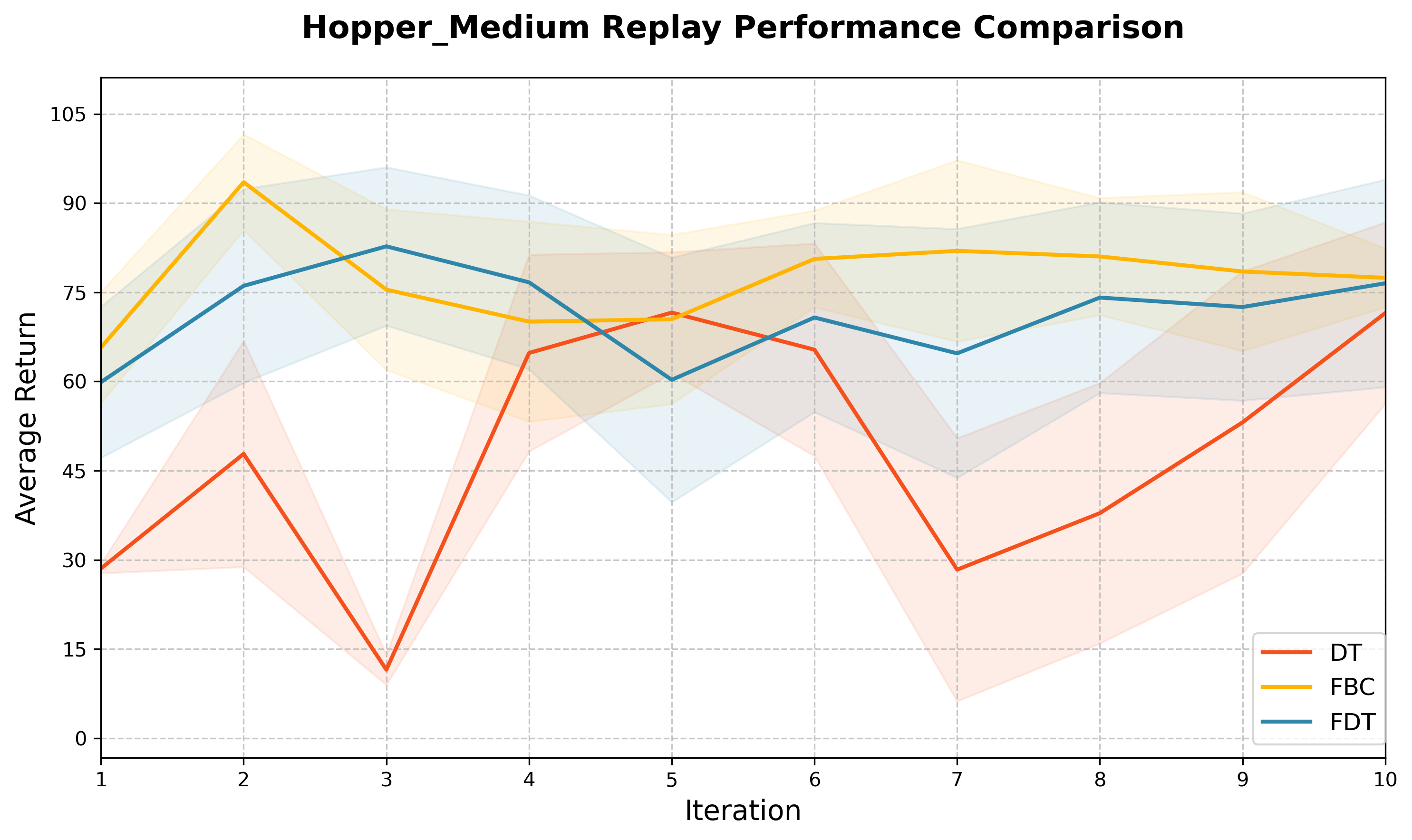}
        \caption{Hopper (Medium-Replay)}
        \label{fig:hopper_medium-replay_comparison}
    \end{subfigure}
    \par
    \vspace{0.5cm}
    \begin{subfigure}[b]{0.32\textwidth}
        \centering
        \includegraphics[width=\textwidth]{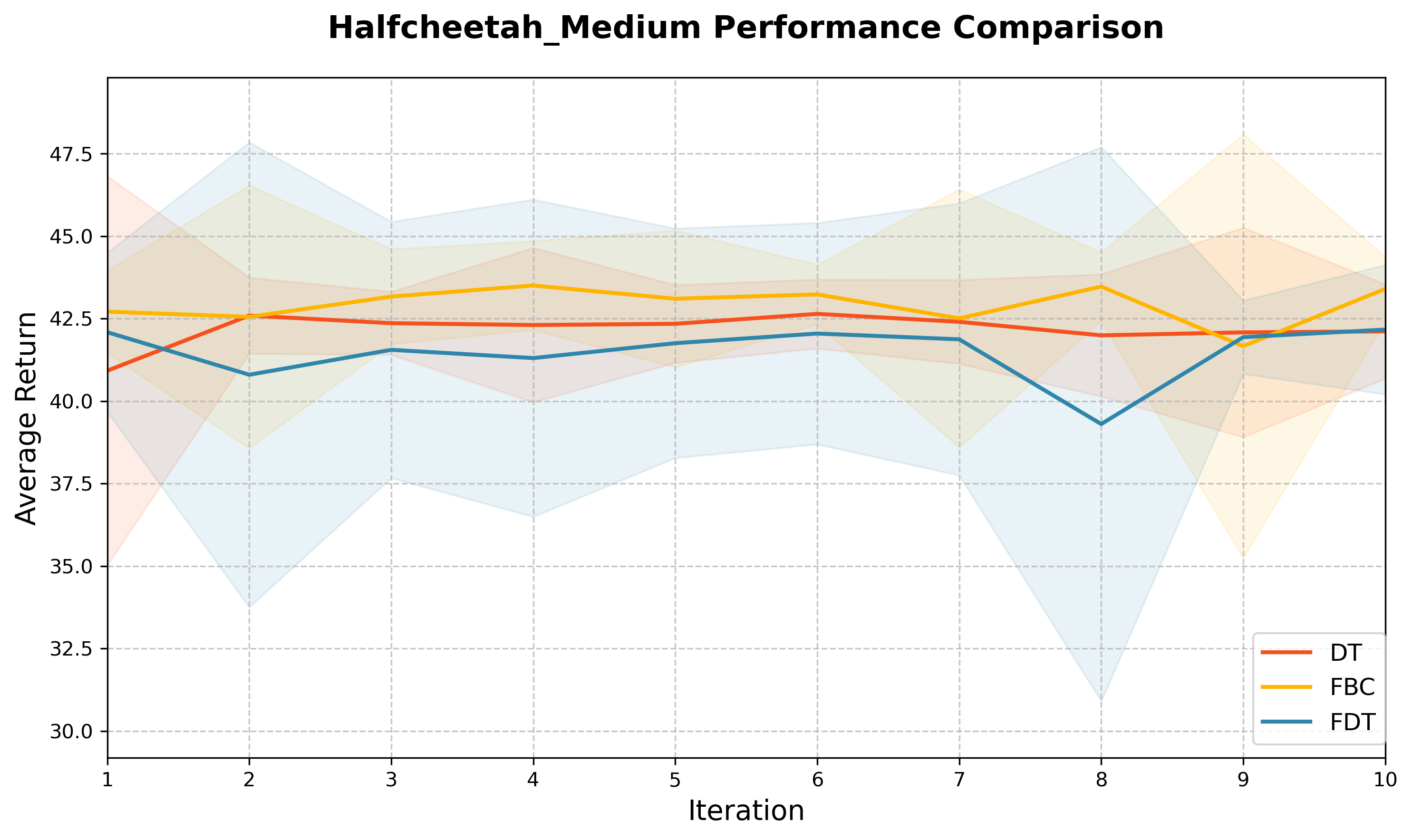}
        \caption{Halfcheetah (Medium)}
        \label{fig:halfcheetah_medium_comparison}
    \end{subfigure}
    \hfill
    \begin{subfigure}[b]{0.32\textwidth}
        \centering
        \includegraphics[width=\textwidth]{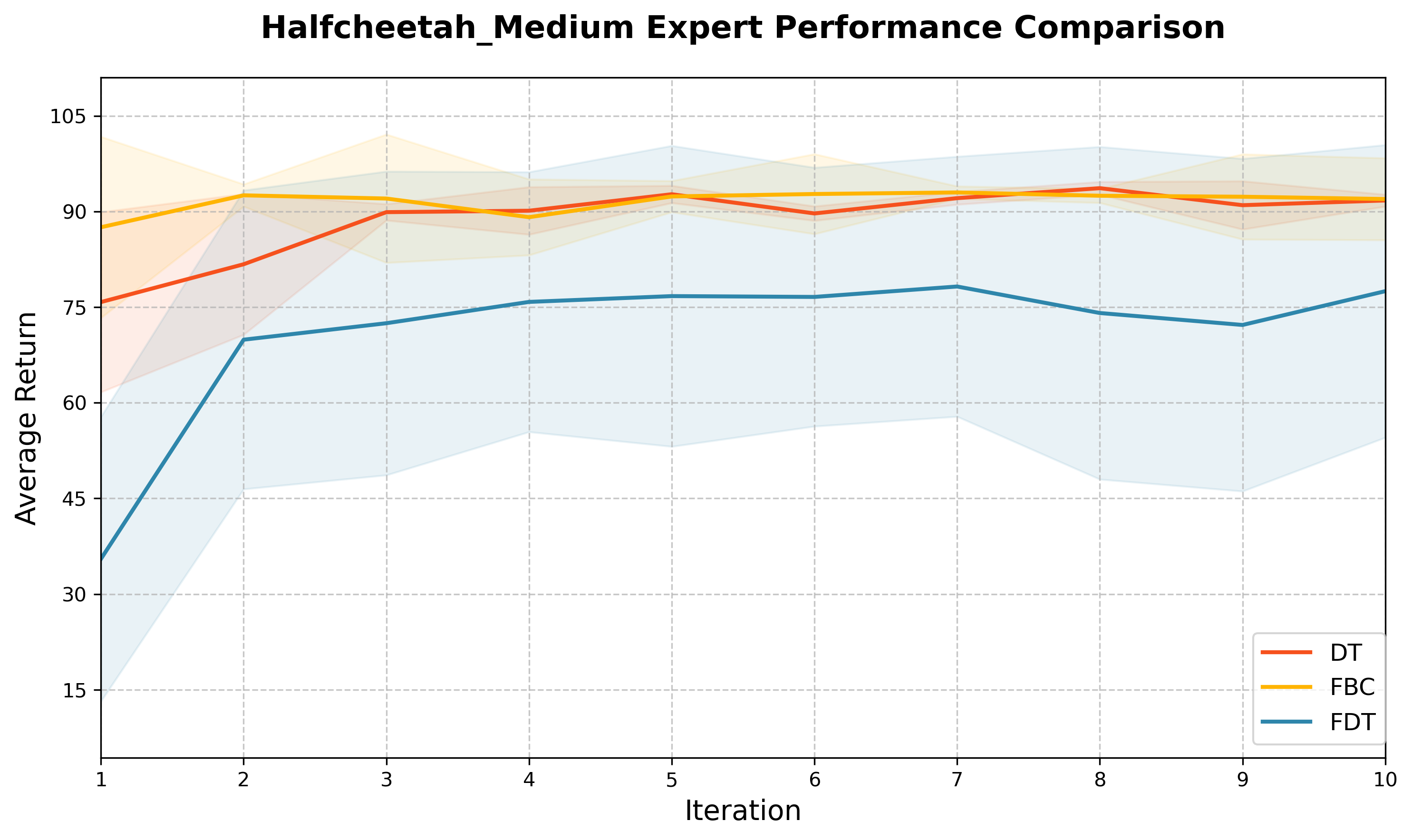}
        \caption{Halfcheetah (Medium-Expert)}
        \label{fig:halfcheetah_medium-expert_comparison}
    \end{subfigure}
    \hfill
    \begin{subfigure}[b]{0.32\textwidth}
        \centering
        \includegraphics[width=\textwidth]{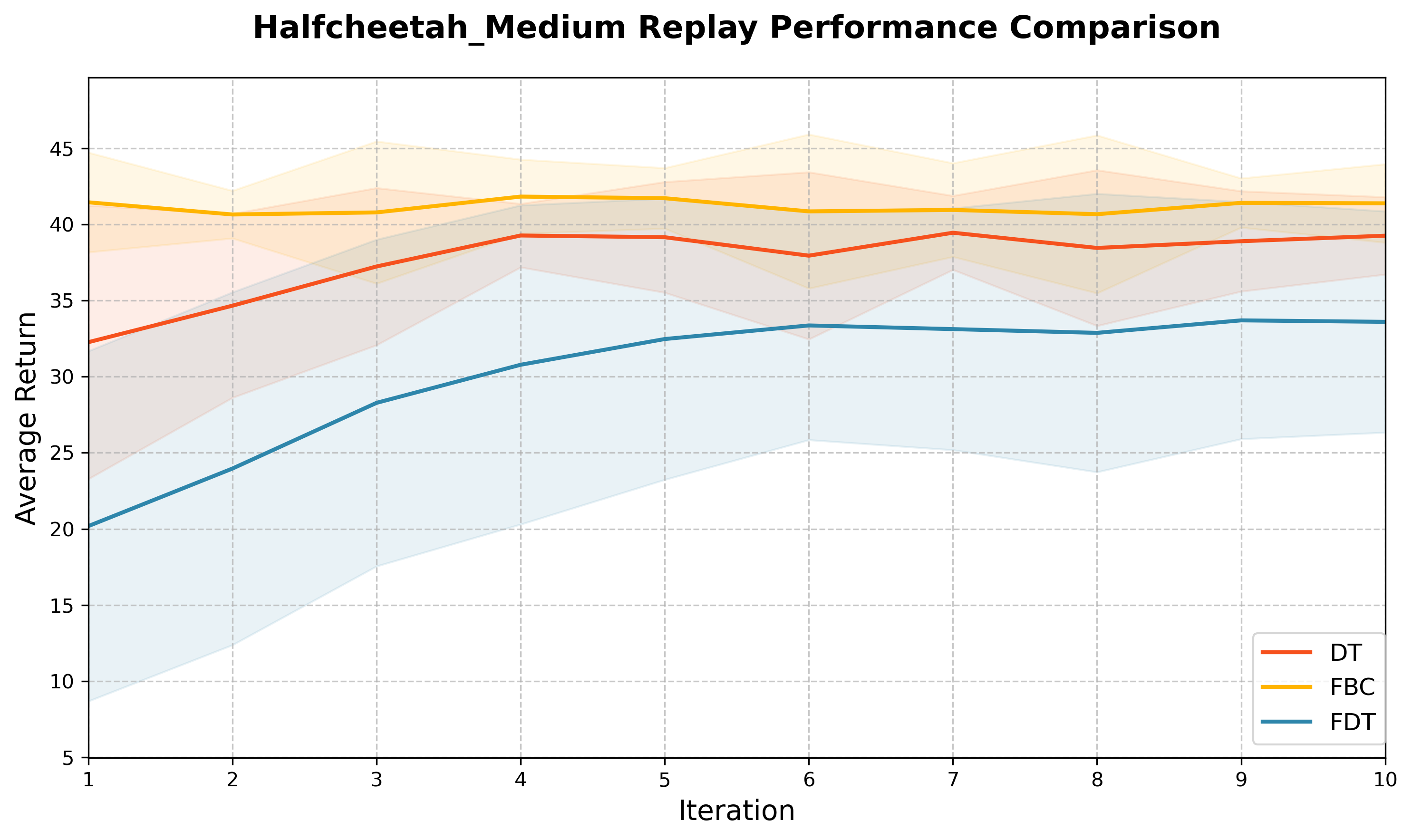}
        \caption{Halfcheetah (Medium-Replay)}
        \label{fig:halfcheetah_medium-replay_comparison}
    \end{subfigure}
    
    \caption{Performance of FBC, FDT, and DT on D4RL.}
    \label{fig:mujoco_sub}
\end{figure}

This section presents the full learning curves for all evaluated methods. Figure~\ref{fig:robomimic_main} shows training performance on the RoboMimic sparse tasks (\textit{Lift} and \textit{PickPlaceCan}), with 95\% confidence intervals. Figure~\ref{fig:mujoco_sub} provides learning curves for the D4RL locomotion benchmarks across all nine task-dataset combinations. CQL curves are excluded for clarity, as comprehensive results are already provided in \citet{Bhargava2024WhenDT} and fall outside the core comparative scope of this study.

% \subsection{Task Descriptions}
% \label{appendix:d4rl-tasks}

% \begin{table}[h]
% \centering
% \caption{Task Descriptions for D4RL Locomotion Environments}
% \label{tab:d4rl-tasks}
% \begin{tabular}{p{3.2cm}p{5.8cm}p{4cm}}
% \toprule
% \textbf{Task} & \textbf{Scene Description} & \textbf{Objective} \\
% \midrule
% \textbf{Hopper} & A planar 2D robot with three joints. & Learn stable forward hopping. \\
% \textbf{Walker2d} & A bipedal robot with six joints. & Learn balanced and stable walking behavior. \\
% \textbf{HalfCheetah} & A planar robot with articulated legs. & Maximize forward velocity while maintaining control. \\
% \bottomrule
% \end{tabular}
% \end{table}

% \subsection{Observation and Action Spaces}
% \label{appendix:d4rl-obs-act}

% The observation and action spaces for all D4RL tasks consist of low-dimensional proprioceptive features, as summarized below.

% \begin{table}[h]
% \centering
% \caption{Observation and Action Space Dimensions for D4RL Tasks}
% \label{tab:d4rl-obs-act}
% \begin{tabular}{ll}
% \toprule
% \textbf{Modality} & \textbf{Dimension (varies by task)} \\
% \midrule
% \textbf{Observations} & \\
% \quad Joint positions and velocities & Typically (17–29,) \\
% \quad Contact and body state info & Included where applicable \\
% \midrule
% \textbf{Actions} & \\
% \quad Joint torque commands & (6–17,) \\
% \bottomrule
% \end{tabular}
% \end{table}

% \subsection{Training and Evaluation Protocol}
% \label{appendix:d4rl-training}

\clearpage
\subsubsection*{Acknowledgments}
\label{sec:ack}
This work is submitted in part by the NYU Abu Dhabi Center for Artificial Intelligence and Robotics, funded by Tamkeen under the Research Institute Award CG010. The experiments were carried out on the High Performance Computing resources at New York University Abu Dhabi.

%%%%%%%%%%%%%%%%%%%%%%%%%%%%%%%%%%%%%%%%%%%%%%%%%%%%%%%%%%%%%%%%
%% NOTE: THIS MARKS THE END OF THE "MAIN TEXT"
%%%%%%%%%%%%%%%%%%%%%%%%%%%%%%%%%%%%%%%%%%%%%%%%%%%%%%%%%%%%%%%%

%%%%%%%%%%%%%%%%%%%%%%%%%%%%%%%%%%%%%%%%%%%%%%%%%%%%%%%%%%%%%%%%
%% Bibliography
%%%%%%%%%%%%%%%%%%%%%%%%%%%%%%%%%%%%%%%%%%%%%%%%%%%%%%%%%%%%%%%%
\bibliography{main}
\bibliographystyle{rlj}

\end{document}